\documentclass[preprint,12pt]{elsarticle}

\usepackage{secdot}
\clearpage
\usepackage{amssymb,epsfig,amsmath,amsthm,amsfonts}

\usepackage[normalem]{ulem}
\usepackage[color,final]{showkeys}
\usepackage{multirow}

\newcommand{\revise}[1]{{\color{blue}#1}}

\journal{European Journal of Operational Research}

\begin{document}

\begin{frontmatter}

\title{Sensitivity Analysis for Computationally Expensive Models using Optimization and Objective-oriented Surrogate Approximations}

\author[ceemath]{Yilun Wang\corref{cor1}}
\ead{yilun.wang@gmail.com}
\author[ceeOrie]{Christine A. Shoemaker\corref{cor2}}
\ead{cas12@cornell.edu}

\address[ceemath]{School of Mathematical Sciences, University of Electronic Science and Technology of China, Chengdu, Sichuan, 611731 China
}
\address[ceeOrie]{School of Civil and Environmental Engineering and School of Operations Research and Industrial Engineering,
Cornell University, Ithaca, New York 14853, USA}


\begin{abstract}
In this paper, we focus on developing efficient sensitivity analysis methods for a computationally expensive objective function $f(x)$ in the case that  the minimization of it has just been performed. Here ``computationally expensive'' means that each of its evaluation takes significant amount of time,  and therefore our  main goal to use a small number of function evaluations of $f(x)$ to further  infer the sensitivity information of these different parameters.  Correspondingly,  we consider the optimization procedure as an adaptive experimental design and re-use its available function evaluations as the  initial design points  to establish a surrogate model $s(x)$ (or called response surface).  The sensitivity analysis is performed on $s(x)$, which is  an lieu of $f(x)$.  Furthermore,   we propose a new local multivariate sensitivity measure, for example, around the optimal solution, for high dimensional problems.  
  Then a corresponding   ``objective-oriented experimental design" is proposed  in order to make the generated surrogate $s(x)$ better suitable for the accurate calculation of the proposed specific local sensitivity  quantities. In addition, we demonstrate the better performance of the Gaussian radial basis function interpolator over Kriging in our cases, which are of relatively high dimensionality and few experimental design points. 
  Numerical experiments demonstrate that  the optimization procedure and the ``objective-oriented experimental design" behavior much better than the classical Latin Hypercube Design.  In addition,  the performance of Kriging is not as good as Gaussian RBF, especially  in the case of high dimensional problems.
\end{abstract}

\begin{keyword}
Sensitivity analysis \sep computationally expensive function  \sep surrogate model  \sep adaptive experimental design \sep global optimization.
\end{keyword}


\end{frontmatter}
\section{Problem Statement, Motivations and Contributions}


In this paper, we are focusing on sensitivity analysis of a black box function $f(x)$ which is defined on a hypercube $\mathcal{D}$ of $\mathbb{R}^n$. $f(x)$ is assumed to be deterministic, continuous, bounded, multimodal and computationally expensive to evaluate, where each of objective function evaluations may take  minutes, hours or even days.
$f(x)$ is expensive to evaluate  typically because it involves high-fidelity computer simulations to study complex, real world physical phenomena,  
in many scientific and engineering fields \cite{Gorissen10SMA} including solutions of systems of partial differential equations. 
For example, in model parameter calibration, $f(x)$ is the discrepancy between  the output of a complex simulation model prediction  $Q^{\mbox{sim}}(x)$ and the observed data and $x \in \mathbb{R}^n$ is the to be estimated parameter vector.

Sensitivity analysis, as a what-if analysis, assesses the contribution of the variation in each input parameter $x_i (i=1, \ldots, n)$ to the variation in the objective function $f(x)$. Our goal is to provide an algorithm that will provide both local and global sensitivity information in a very computationally efficient fashion for black box computationally expensive multimodal function for which derivative information is not available. In particular, the algorithm can provide accurate solution for the follows:
  \begin{enumerate}
       \item global sensitivity results based on a specific  method such as  ``Extended FAST" method.
       \item local sensitivity: 
  numerical univariate derivatives $\frac{f(x^f+\Delta x)-f(x^f)}{\Delta x}$ for  $\Delta x =\{\Delta x_1, \Delta x_2, \ldots, \Delta x_n\}$ of variable magnitudes,  for any point $x^f\in \mathbb{R}^n$ in the domain $\mathcal{D}$, and our proposed local multivariate sensitivity quantities around given $x^f$.

    \end{enumerate}
 Here, sensitivity analysis is performed after searching for the global minimum of $f(x)$. 
Typically one would use $x^f=x^*$ when considering local sensitivity, where $x^*$ is the searched global minimum.   We make use of surrogate approximation and optimization to achieve computational efficiency of sensitivity analysis.

 Traditionally, a simple sensitivity analysis of $f(x)$ is often acting as a prerequisite  of
 optimization of $f(x)$ by  screening out the very insensitive parameters, especially when $n$ is very big,
 for example, $n$ is hundreds.
On the contrary, in this paper, we  consider the situations where ones want to performs sophisticated sensitivity analysis on the remaining parameters of $f(x)$, after the above parameter screening. For example, in the field of parameter calibration, Sorooshian and Arfi studied the importance and meaning of the sensitivity analysis for the post-calibration studies \cite{Sorooshian82SA}.  In this paper, we will 
focus more on  how to efficiently  obtain the sensitivity information of the computationally expensive function $f(x)$, based on a very limited or affordable number of function evaluation of it. 

One way to overcome  computational difficulty is to establish a  surrogate model $s(x)$ (also called response surface, metamodel) as an approximation  of $f(x)$ based on an affordable number of function evaluations of $f(x)$, and then perform the function-evaluation-intensive  sensitivity analysis on $s(x)$.  One key point for the establishment of the surrogate surface  is to properly pick the locations of the evaluation points of $f(x)$, which are also called experimental design points in statistics literature \cite{Kleijnen05DOESA}.  In this paper, we focus on how to obtain these experimental design points, in order to make the generated response surface of better approximation accuracy for the calculation of either global sensitivity  and local sensitivity quantities. In addition, we compared the performance of two typical surrogate surfaces, Radial Basis Function (RBF) interpolator  and Kriging.

\subsection{Relationship with Existing Work} \label{sec:ExistingWork}


The existing work of establishing a surrogate of a computationally expensive continuous black box function $f(x)$ is mainly about
global sensitivity analysis, because the traditional linear local sensitivity analysis methods do not   
 requires a huge  number of
 function evaluations.  Though we have the limited computational budget on the number of function evaluations of $f(x)$, we do have the freedom to determine where to perform these function evaluations and what kind of surrogate is adopted to approximate $f(x)$.
 The process of determining the positions to perform the function evaluations is called ``Experimental Design'' and the determined positions are called design points.   The experimental design is often related with the choice of surrogate models, which typically include polynomial response surface, rationally functions, splines, Support Vector Machines, Kriging/Gaussian Processes (GP), radial basis function interpolators or other Artificial Neural Networks (ANN) (Buhmann, 2003).
%
%
%
Here we will briefly review some of corresponding state of the art experimental design methods \cite{Kleijnen05DOESA,Kleijnen08SimulationExp,Kleijnen08BookDOE,Kleijnen09Review,Queipo2005SurrogateSAandOP,Jin00comparativestudies,Simpson01Survey,Shan10SurveyHEB}. 

For complex surrogate models, such as Kriging and artificial neural networks, it is generally believed that space-filling stochastic designs are suitable for them, because these designs try to specify the design points so that as much of the design space is sampled as possible within the allowed maximal number of function evaluation points.  Space-filling designs typically include  Latin hypercube design \cite{McKay79LHD}, various optimal or orthogonal designs including  Minimax and Maximin design \cite{Johnson90Maximin}, Entropy Design \cite{Shewry97Entropy,Currin91EntropyDesign} and orthogonal arrays \cite{Owen92OA}.  
As we have seen, the above mentioned space filling designs do not depend on the specific underlying simulation model \cite{Crary02MetaModel}, and  all design points are simultaneously optimized according to one of the above criteria.  They are often called ``the one-stage designs'' \cite{Sacks1989DACE}. SUMO--Surrogate MOdeling Lab \cite{Gorissen10SMA} has implemented most of the above surrogate models and experimental designs.
However, because $f(x)$ behaviors like a black box and its evaluation is computationally expensive, the shape of $f(x)$ and the optimal distribution of design points are not known up front and therefore we can not guarantee the above ``evenly'' distributions are best for every $f(x)$.

Contrary to one stage designs,  sequentialized designs  are expected to be more efficient in terms of requiring fewer function evaluations of $f(x)$ to establish a faithful surrogate model $s(x)$ \cite{Sacks1989DACE,Park02sequential}.  Sequential designs imply that the underlying function is better analyzed often via a surrogate model established
based the function evaluations at the previous design points  before determining the next design point, i.e. the design is customized for different specific underlying models.  So the sequentialized design is often also called active learning, or adaptive sampling in statistical literature.
Different sequentialized designs might focus on catching different features of the underlying function $f(x)$.
Some recent efforts along these directions including Kriging based sequentialized designs by Kleijnen et al.  \cite{Kleijen04DeterministicSequentialDesign,Kleijnen09Review}.  These works were based on the improved Kriging variance formulas via  the bootstrapping \cite{Hertog06CorrectVariance,vanBeers2008Kriging} technique or cross-validation and jackknifing techniques \cite{Kleijnen09Review}, rather than the classic formula used in the literature e.g.  \cite{Cressie93Kriging} and  \cite{Sacks1989DACE,Jones98EGO}, because the classic Kriging variance formula neglects estimation of  certain correlation parameters of Kriging, which makes the Kriging predictor a nonlinear estimator  \cite{Hertog06CorrectVariance,Kleijnen10KrigingOPT}, and  therefore the classic one  is expected to underestimate the true variance \cite{Hertog06CorrectVariance,Cressie91Book}.
As for the radial basis function based emulator, Jin et al. \cite{Jin02SequentialSampling} presented an approach based on cross-validation; 
Shan \cite{shan10HEB, Shan10SurveyHEB} proposed a sequentialized design for RBF
surface called RBF-HDMR which integrates the radial basis function with a high dimensional
model representation first proposed by Sobol \cite{Sobol93SA} and currently could not be applied to functions where its parameters have highly nonlinear interactions.
However, Jin at al \cite{Jin02SequentialSampling} found that many of the current sequential sampling approaches were not necessarily better than
one-stage approaches such as the optimal LHD, partially because the information based on the early created surrogate models might be misleading or incomplete, or not properly adopted. Therefore, the LHD or the optimal LHD is still widely used in the practical applications.
In terms of sequential designs, we point out that
most optimization algorithms belong to this family since they determine the next function evaluation point based on the knowledge of $f(x)$ provided by the previous function evaluation points, and their   
function evaluations can  often effectively infer the major global trends of $f(x)$.  

As for local sensitivity analysis, it mainly provides the slope of the model output in the parameter space at a nominal point or called a base case $\bar{x}$. For example, in the field of parameter calibration of simulation model, local sensitivity analysis is able to provide some quantitative idea about the shape of the objective function in the vicinity of the estimated parameters  obtained by the calibration procedure . Specifically, it helps us establish some measure of confidence
%
%
in the parameters and hence the fitting criterion
employed in the definition of the objective function, and detect non-identifiability of parameters,
leading to appropriate modification of the simulation model \cite{Sorooshian82SA}. Traditionally, they are mostly based on gradients or numerical approximation of gradients at the nominal point,  usually requires a small number of function evaluations, and therefore typically do not need the incorporation of surrogates.
 However, gradients only provide the information within a small vicinity of the nominal point for nonlinear functions and fail to take the parameter interactions into considerations. Therefore we aim to propose a new local sensitivity analysis method which takes nonlinearity and parameter interactions into consideration. Correspondingly, the required number of function evaluations increases a lot and is often even beyond the allowed computational budget. In such cases, we also turn to the help of the surrogate as the global sensitivity analysis does, and
%
 we would like to develop a tailored experimental design for this new local sensitivity analysis method with aim to reduce the size of the experimental design by giving up the global approximation property of $s(x)$ and only focusing on the local approximation tailored for calculating 
the  local sensitivity quantities.  
\subsection{Our Contributions}
Our first contribution is to  build a bridge between the optimization and the following sensitivity analysis, from the computational point of view,  via the adoption of surrogate surfaces.  Specifically, the function evaluations during optimization are not discarded. Instead,  we saved and reused them for the establishment of
%
a surrogate surface $s(x)$ of $f(x)$, i.e.  the function evaluation points during the optimization are the initial experimental design points. 
$s(x)$ is acting as
an  approximation of $f(x)$ for the following sensitivity analysis, because its evaluation is computationally cheap. More important, we show that  the
optimization step,  as an active experimental design, 
outperforms  some classical  non-adaptive
experimental designs. 

%

 The second contribution is to  propose a new local sensitivity analysis method (for example, around the optimal solution), and present a corresponding tailored experimental design to efficiently generate a surrogate of good local approximation  property and suitable for the calculation of its sensitivity quantities. 
 Unlike many traditional local sensitivity analysis methods which assume that $f(x)$ is nearly linear and do not take the parameter interactions into consideration, this new one  ranks the  parameters based on simultaneous perturbations of several parameters around the nominal point and therefore takes the nonlinearity and parameter interactions into considerations. In addition, it might consider a much larger perturbation step size than traditional local sensitivity measures.

 For the high dimensional
 functions, 
 the required function evaluations might still be unaffordable, though it typically required much less function evaluations than most global sensitivity analysis methods. In order to circumstance the computational difficulty, we also turn to the establishment of a surrogate model $s(x)$ as a lieu of the original function $f(x)$. However, unlike the traditional experiment design methods which  are usually devised for a surrogate of a good global approximation property,  we propose an idea of ``objective-oriented " experimental design  method, which is tailored only for better accuracy of the calculated local sensitivity measures on the generated response surface, instead of pursuit of the more strict global approximation accuracy,  in order to significantly reduce the number of the required experimental design points. 

The third contribution is that besides considering the issue of experimental design methods, we compare the performance  of different surrogate types, especially  the Gaussian RBF and Kriging in both relatively low and high dimensional problems.

The fourth contribution is that we are  evaluating  different experimental design methods through the accuracy of the calculated sensitivity quantities, rather than the approximation errors of the resulted surrogates as many existing works do,
since the ultimate goal of establishing a surrogate is to efficiently calculate the sensitivity quantities here. 
In addition, 
we will also be considering the relatively high dimensional space, rather than very low dimensional problems ($n \le 5$), which are main targets considered in most previous work. 


\subsection{Paper Organization}
The following part of this paper consists of $4$ sections. 
Section $2$ gives a detailed description of our methodology, and the new local multivariate sensitivity analysis quantities.  
Section $3$ demonstrates the advantages of our experimental design scheme over other alternatives through both synthetic problems and real application examples. We also compare the performance of RBF interpolator and Kriging.   
The summary and future work are given in Section $4$.

\section{Our Methodology} \label{Sec:Methodology}
%
We propose to
reuse the available function evaluations during the optimization procedure as the initial design points for establishment of the surrogate for the following sensitivity analysis, no only because they are ``free'' in terms of sensitivity analysis, but also they indeed greatly improve the global approximation accuracy  of
 the generated surrogate, as an adaptive experimental design. 
 Next, we further
extend the set of the design points by adopting other appropriate experimental design methods in order to further improve approximation property of the obtained surrogate. This choice of the extended design points are related with what kind sensitivity analysis is performed, for example, local
sensitivity analysis or global sensitivity analysis. We propose a ``objective-oriented" experimental design to produce a surrogate surface of a better approximation quality. 
We denote our sensitivity analysis framework as  O3AED (Optimization and Objective-Oriented Adaptive Experimental Design for surrogates assisted sensitivity analysis).  In the following parts, we first give a brief introduction to the framework, and  
then introduce our new local sensitivity measure around the optimal solution and its corresponding tailored experimental design method.  

\subsection{Algorithmic Framework of O3AED  and Our Main Contributions}

An important feature of O3AED  is to consider optimization and sensitivity analysis into an integrated framework.  Optimization  not only returns the optimal solution which the following local sensitivity analysis is performed around, but also provides its function evaluations for helping generate a surrogate on which the following sensitivity analysis is performed.  
%
%
Simply, O3AED  consists of several steps listed as follows:  

\begin{itemize}

\vspace{-0mm}

\item[] {\bf Step 1:} Search for the minimum of  the objective function $f(x)$ using a global optimization algorithm, save all the executed function evaluations, and set these function evaluation points  as the initial design points.

\item[] {\bf Step 2:} Add more design points, where evaluations of $f(x)$ will performed in order to obtain a more accurate surrogate. Depending on global sensitivity analysis or local sensitivity analysis, 
     The ways of adding new design points may vary and be adaptive. 
 \item[] {\bf Step 3:} Construct  a surrogate $s(x)$ based on the above  evaluations of $f(x)$.
\item[] {\bf Step 4:} Perform sensitivity analysis on the surrogate $s(x)$ in place of the original computationally expensive objective function $f(x)$.
\end{itemize}

Before  moving to detailed explanation of each step, we first introduce  the sensitivity analysis methods used in our paper, since the implementations of  Step $2$ and Step $3$ are also based on the specific sensitivity analysis methods.
We first review the global sensitivity analysis method used in our paper. 
Then we introduce the motivation and the definition of our new local multivariate sensitivity measure, as well as its corresponding innovative experimental design method. 

\subsection{Brief Review of Extended FAST}\label{subsec:EFAST}
%


%
There have existed many kinds of global sensitivity analysis methods, which might be suitable for different kinds of underlying functions. As for nonlinear and non-monotonic
relationships between model inputs and  outputs, the variance based methods
include Sobol' method \cite{Sobol93SA}, classic Fourier amplitude sensitivity test (FAST) and the extended FAST \cite{Saltelli99ExtendedFAST} are widely used.
In this paper, we take the extended FAST as the example, though other methods could be applied here. 
It provides a measure of fractional variance accounted for by individual variables.
For each variables, Extended FAST returns two kinds of sensitivity quantities, i.e. the first order sensitivity index $S_i$ and total sensitivity index $ST_i$ where $S_i$
%
%
%
%
measures the main effect of $x_i$ on the output variance and 
%
$ST_i$ also considers the parameter interactions and is the proportion of contribution of $x_i$ to the total variance of outputs. If input variables have no internal interactions, we have $ST_i = S_i$. Otherwise, $ST_i > S_i$.  $ST_i$ represents the contribution of the input variable $x_i$ to the variance of the objective function $f(x)$.  The bigger $ST_i$, the more sensitive $x_i$ is.



The calculation of $ST_i$ and $S_i$ is mainly composed of high dimensional integrals.  In practice, the analytic formula for them are not available due to the ``black box'' feature of $f(x)$ and their evaluation is often through  Monte Carlo sampling, which relies on repeated random sampling to compute their results.  When the problem is a high dimensional problem,
a very large number of samples might be  required. Therefore, even for the non-computationally expensive functions,  their dimensions can not be very high, in order to make Extended FAST computationally feasible on common personal computers. For relatively low dimensional problems, Extended FAST is a very efficient method compared with many other alternatives. However, the required function evaluations of $f(x)$ might be still unaffordable when $f(x)$ is a computationally expensive function. In such cases, a surrogate might be adopted  and we will compare performance of our method with that of other alternatives of experimental designs for establishing a proper surrogate model.

\subsection{New  Multivariate Local Sensitivity Measures and Corresponding Tailored Experimental Design}\label{subsec:MVMLS}




Most  traditional local sensitivity analysis methods are executed by varying input parameters one-at-a-time by a very small perturbation. Let $\rho$ be a fixed percentage  of the range of each coordinate and the corresponding step size is
$\Delta=\rho\times(b-a)$.   The traditional univariate perturbation adopts a small $\rho$ and the corresponding elementary effect is defined as follows.
\begin{equation}
\bar{x}^{(k^{+},\rho)}=[
\bar{x}_1, \ldots, \bar{x}_{k-1}, \bar{x}_k+\Delta, \bar{x}_{k+1}, \ldots, \bar{x}_n ]
\end{equation}

\begin{equation}
\bar{x}^{(k^{-},\rho)}=[
\bar{x}_1, \ldots, \bar{x}_{k-1}, \bar{x}_k-\Delta, \bar{x}_{k+1}, \ldots, \bar{x}_n ]
\end{equation}

\begin{equation}\label{Def:SIkR}
SI_{k^{+}}^{1,\rho}=\left|
\frac{f(\bar{x})-f(\bar{x}^{(k^{+},\rho)})}{f(\bar{x})}\right|, \quad  \quad SI_{k^{-}}^{1,\rho}=\left|
\frac{f(\bar{x})-f(\bar{x}^{(k^{-},\rho)})}{f(\bar{x})}\right|
\end{equation}

In order to take the function nonlinearity into consideration, we propose to use multiple large values of $\rho$, for example $\rho=0.1, 0.2, \ldots, 0.4$.

We also expect it to help identify the nature of the parameter interactions .

Next, we also need to consider the parameter interactions and  verify that they conform with our understanding
of the true processes involved in order to  detect suboptimal solutions \cite{Sorooshian82SA}. Correspondingly we perturb multiple parameters simultaneously, and
%
 we consider up to  simultaneous $3$-parameter perturbation in this paper.
If two-at-a-time (TAT) perturbation  is performed,  the following $4$ bivariate perturbation samples are required to calculate the  corresponding elementary effects.

\begin{equation}
\end{equation}
\begin{eqnarray}\nonumber
\bar{x}^{(k^{+},j^{+},\rho)}&=&[
\bar{x}_1, \ldots, \bar{x}_{i-1}, \bar{x}_i+\Delta, \bar{x}_{i+1}, \ldots, \bar{x}_{j-1}, \bar{x}_j+\Delta, \bar{x}_{j+1}, \ldots,\bar{x}_n ]\\\nonumber
\bar{x}^{(i^{+},j^{-},\rho)}&=&[
\bar{x}_1, \ldots, \bar{x}_{i-1}, \bar{x}_i+\Delta, \bar{x}_{i+1}, \ldots, \bar{x}_{j-1}, \bar{x}_j-\Delta, \bar{x}_{j+1}, \ldots,\bar{x}_n ]\\\nonumber
\bar{x}^{(i^{-},j^{+},\rho)}&=&[
\bar{x}_1, \ldots, \bar{x}_{i-1}, \bar{x}_i-\Delta, \bar{x}_{i+1}, \ldots, \bar{x}_{j-1}, \bar{x}_j+\Delta, \bar{x}_{j+1}, \ldots,\bar{x}_n ]
\\\nonumber \bar{x}^{(i^{-},j^{-},\rho)}&=&[
\bar{x}_1, \ldots, \bar{x}_{i-1}, \bar{x}_i-\Delta, \bar{x}_{i+1}, \ldots, \bar{x}_{j-1}, \bar{x}_j-\Delta, \bar{x}_{j+1}, \ldots,\bar{x}_n ]
\end{eqnarray}

\begin{equation}\label{Def:SIkRR}
SI_{(k^{+},j^+)}^{2,\rho}=\left|
\frac{f(\bar{x})-f(\bar{x}^{(k^{+},j^{+},\rho)})}{f(\bar{x})}\right|
\end{equation}
$(\bar{x}^{(k^{+},j^{-},\rho)},  SI_{(k^{+},j^-)}^{2,\rho})$,  $(\bar{x}^{(k^{-},j^{+},\rho)},  SI_{(k^{-},j^+)}^{2,\rho})$, $(\bar{x}^{(k^{-},j^{-},\rho)},  SI_{(k^{-},j^-)}^{2,\rho})$ are defined in a similar way.


If three-at-a-time (THAT) perturbation  is performed,  the following $8$ trivariate perturbation samples are required to calculate the  corresponding elementary effects.


\begin{equation}
\bar{x}^{(k^{+},j^{+},i^{+},\rho)}=[\bar{x}_1, \ldots,  \bar{x}_k+\Delta,  \ldots, \bar{x}_{j}+\Delta, \ldots, \bar{x}_{i}+\Delta, \ldots, \bar{x}_n]
\end{equation}

\begin{equation}\label{Def:SIkRRR}
SI_{(k^{+},j^+,i^+)}^{3,\rho}=\left|
\frac{(f(\bar{x})-f(\bar{x}^{(k^{+},j^{+},i^{+},\rho)})}{f(\bar{x})}\right|
\end{equation}
$(\bar{x}^{(k^{+},j^{+},i^{-},\rho)}, SI_{(k^{+},j^+,i^-)}^{3,\rho})$, $(\bar{x}^{(k^{+},j^{-},i^{+},\rho)}, SI_{(k^{+},j^-,i^+)}^{3,\rho})$, $(\bar{x}^{(k^{+},j^{-},i^{-},\rho)}, SI_{(k^{+},j^-,i^-)}^{3,\rho})$, \\ $(\bar{x}^{(k^{-},j^{+},i^{+},\rho)}, SI_{(k^{-},j^+,i^+)}^{3,\rho})$, $(\bar{x}^{(k^{-},j^{+},i^{-},\rho)}, SI_{(k^{-},j^+,i^-)}^{3,\rho})$, $(\bar{x}^{(k^{-},j^{-},i^{+},\rho)}, SI_{(k^{-},j^-,i^+)}^{3,\rho})$, \\ $(\bar{x}^{(k^{-},j^{-},i^{-},\rho)}, SI_{(k^{-},j^-,i^-)}^{3,\rho})$ are defined in a similar way.


In summary,   we consider from univariate perturbations to  multivariate perturbations on various  perturbing  sizes,  in order to account for the parameter interactions and nonlinearity.  We call it as MultiVariate Multi-Steps Local sensitivity analysis method (MVMSL, for short).

Let $n$ be the number of parameters, and the numbers of univariate perturbation samples, bivariate perturbation samples, trivariate perturbation samples  are $2n$, $2n(n-1)$, $4n(n-1)(n-2)/3$, respectively, for a given $\rho$. 
Therefore, if $n$ is big,  this number of function evaluations might still not be affordable, especially when $f$ is computationally expensive and ones might try multiple $\rho$ values (for example, $\rho=0.1, 0.2, 0.3, 0.4$).
In such cases, we will also turn to the surrogate model with aim to reduce the function evaluations of $f(x)$.  In order to establish a suitable surrogate for the calculation of  the quantities $f(\bar{x}_1, \ldots, \bar{x}_{k-1}, \bar{x}_k+\Delta, x_{k+1}, \ldots, x_n), f(\bar{x}_1, \ldots,  \bar{x}_k+\Delta,  \ldots, \bar{x}_{j}+\Delta, \ldots, x_n), f(\bar{x}_1, \ldots,  \bar{x}_k+\Delta,  \ldots, \bar{x}_{j}+\Delta, \ldots, \bar{x}_{i}+\Delta, \ldots, x_n)$ and etc.,   we propose to pick the allowed number of experimental design points from the set of all the univariate perturbation samples , bivariate perturbation samples and trivariate perturbation samples in a random way. 
A detailed explanation is in Section \ref{subsec:MVMLSDesign}.

As for bivariate perturbations and trivariate perturbations, one might be interested in the most few sensitive duos or triples by sorting the corresponding element effects defined as (\ref{Def:SIkRR}) or (\ref{Def:SIkRRR}). Meanwhile, one might be also interested in the ranks of the input variables $x_i$ $(i=1, 2, \ldots, n),$  in term of bivariate perturbations and trivariate perturbations. In the following section \ref{subsec:MVMLSSA} we introduce a simple statistical way to rank the parameters. In section \ref{subsec:SVD}, another ranking method through the eigenvalue decomposition of the hessian matrix is also reviewed. 

%


\subsubsection{The First Way to Define Local Sensitivity Indices} \label{subsec:MVMLSSA}

Given the element effects defined, for example,  by (\ref{Def:SIkR}), (\ref{Def:SIkRR}), (\ref{Def:SIkRRR}),  we calculate the average for each combination as follows:
\begin{eqnarray*}\label{Def:SIk}
SI_{i}^{1,\rho}&=&
(SI_{i^{+}}^{1,\rho} +
SI_{i^{-}}^{1,\rho})/2\\
SI_{i,j}^{2,\rho}&=&([
SI_{(i^{+},j^+)}^{2,\rho}+
SI_{(i^{+},j^-)}^{2,\rho}+
SI_{(i^{-},j^+)}^{2,\rho}+
SI_{(i^{-},j^-)}^{2,\rho}])/4\\
 SI_{i,j,k}^{3,\rho}&=&([
SI_{(i^{+},j^+,k^+)}^{3,\rho}+
 SI_{(i^{+},j^+,k^-)}^{3,\rho}+
 SI_{(i^{+},j^-,k^+)}^{3,\rho}+
 SI_{(i^{+},j^-,k^-)}^{3,\rho}\\&+&
 SI_{(i^{-},j^+,k^+)}^{3,\rho}+
SI_{(i^{-},j^+,k^-)}^{3,\rho}+
 SI_{(i^{-},j^-,k^+)}^{3,\rho}+
 SI_{(i^{-},j^-,k^-)}^{3,\rho}])/8
\end{eqnarray*}

For each parameter $x_k$, we can also calculate its several 
sensitivity quantities based on bivariate perturbations and trivariate perturbations, respectively.

\begin{eqnarray}\label{Def:SIk2}
SI_{i}^{2, \rho}&=&  \frac{\sum_{j=1, j\neq i}^n SI_{(i,j)}^{2,\rho}}{n-1}\\\label{Def:SIk3}
SI_{i}^{3, \rho}&=&   \frac{\sum_{j=1, j\neq i}^n\sum_{l=j+1}^{n} SI_{(i,j,l)}^{3,\rho}}{(n-1)\times(n-2)/2}
\end{eqnarray}

In summary, $SI_{i}^{1,\rho}, SI_{i}^{2,\rho}, SI_{i}^{3,\rho}$ are $3$ sensitivity quantities of the input variable $x_i (i=1, 2, \ldots, n)$, in terms of
univariate perturbations, bivariate perturbations, and trivariate perturbations, respectively, for a given $\rho$. The larger the sensitivity quantity, the more sensitive the corresponding input variable is.

\subsubsection{The Second Way to Define Local Sensitivity Indices} \label{subsec:SVD}
We have another way to evaluate the sensitivity ranks of the parameters by simultaneously consider the univariate perturbations and bivariate perturbations, i.e.  $SI_{i}^{1, \rho}$ and $SI_{i,j}^{2,\rho}$ $(i=1,\ldots, n, j=1,\ldots, n)$, as follows.

%


Given the perturbation size $\rho$, we can have the following matrix
univariate perturbations and bivariate perturbations. 

\begin{eqnarray*} H^{\rho}=
\left(
  \begin{array}{cccc}
     SI^{1,\rho}_{1} &SI^{2,\rho}_{1,2}& \ldots & SI^{2,\rho}_{1,n} \\
    SI^{2,\rho}_{2,1} &SI^{1,\rho}_{2} & \ldots & SI^{2,\rho}_{2,n} \\
    \vdots & \vdots & \ddots & \vdots \\
    SI^{2,\rho}_{n,1} & SI^{2,\rho}_{n,2} & \ldots & SI^{1,\rho}_{n} \\
  \end{array}
\right)
\end{eqnarray*}

An eigenvalue decomposition is performed on $H^{\rho}$. The eigenvectors corresponding to eigenvalues of large absolute value are the directions of large curvatures. In this paper, we consider the eigenvectors $U^{1}$ and $U^{2}$, which are corresponding to the two eigenvalues of the biggest absolute value.  For each input variable $x_i$, its corresponding sensitivity quantities based on eigenvalue decompositions are $SI_{i}^{E,1, \rho} \doteq |U^{1}_i|$ and $SI_{i}^{E,2, \rho}\doteq|U^{2}_i|$, where $|\cdot|$ represents the absolute value.    The larger $SI_{i}^{E,1, \rho} $ or $SI_{i}^{E,2, \rho}$ is, the more sensitive $x_i$ is.




\subsubsection{Objective-Oriented Adaptive Experimental Design Methods} \label{subsec:MVMLSDesign}

 For the local sensitivity indices based on the multi-variate perturbation, the number of the required function evaluations can be still too large for computationally expensive functions, especially those of high dimension, even though it is already  computationally much cheaper than most of global sensitivity analysis methods. 
 Correspondingly, we also turn to the help of the surrogate, 
 and the key point is still about how to establish a good surrogate which is well approximating the true function $f(x)$ for this specific purpose. 
%
%
%
%
%
%

For $f(x)$ defined in a relatively high dimensional space with complicated input-output relationship, it is hard to get a surrogate of global approximation based on an affordable number of function evaluations. In such cases, establishing a global approximation of high fidelity might be a waste and unrealistic no matter for the one-time space filling designs or sequentialized designs are adopted.
  Correspondingly we proposed to 
develop a specific design to generate a surrogate which  may be of great local approximation property and only  suitable for the calculation of the function values at those MVMSL samples. In order to make the generated surrogate have a good approximation to the true function  $f(x)$ at the samples, we try to make the set of the evaluation points (or called experimental design points) close to the set of samples. It is very important, especially when the perturbation step is large and a big vicinity of the prescribed point is considered.  A easy way is to randomly pick a subset of the MVMSL samples as the experimental design points to generate the surrogate surface. For example, in this paper, we can  choose all the one-variate perturbation samples, a randomly picked small portion of two-variate perturbation samples and three-variate perturbation samples, as the set of the experimental design points, and this kind of ``constrained randomness"  is very effective, demonstrated by numerical experiments in Section \ref{Sec:NumExp}. 


\subsection{Detailed Algorithmic Description}
\subsubsection{Step 1: Initialization by  Optimization} \label{Sec:Review_OPT}

     The optimization mainly aims to  (1) find the optimal solution of $f(x)$, around which  a local sensitivity analysis might be performed later; (2)  
provide its function evaluations  for the generation of a surrogate model of global approximation property where sensitivity analysis  can be performed on. %


There have been many algorithms for minimizing computationally expensive functions with box constraints \cite{Rios09comparision}  including scattering search, dynamically dimensioned search, simulated annealing, genetic algorithms, multi-start frameworks for local optimization such as OQNLP, and direct search methods \cite{Gutmann01RBF,Jones98EGO,VandenBerghen2005Condor,Tolson07DDS,Regis2007PRBF,Kleijnen10KrigingOPT,Regis11Constraints},  response surface based evolution algorithm \cite{Ong03SurrogateEA,Ong08HERBF,Jin05Comparison} and pattern search algorithms \cite{Dennis97KrigingPatternSearch}, and trust region algorithm \cite{Brekelmans05DFO,Powell08NEWUOA}.  Other popular metaheuristic optimizers include particle swarm optimization \cite{Kennedy95PSO}, differential evolution \cite{Storn97DES} and etc.
%
According to the No Free Lunch (NFL) theorems (Wolpert and Macready 1997) \cite{Wolpert97NFL}, different methods may be fitful for different kinds of problems and no single method can all perform the best in general in terms of finding the optimal solution using a small number of function evaluations of $f(x)$.  In general, most optimization algorithms can be considered as adaptive experimental designs and their performs function evaluations could reflect the global shape of $f(x)$ to certain degree. 

\subsubsection{Step 2: Add More Design Points}
 Step $2$ aims to expand the set of design points  initialized by the optimization step, in order to generate  a more faithful  surrogate model $s(x)$ for the following sensitivity analysis.  Notice that Step $2$ and Step $3$ and Step $4$ are in fact closely related with each other and should be considered together. Depending on different sensitivity analysis method in Step $4$, ones might adopt corresponding experimental design methods to extended the  set of design points. In the paper, we
 consider both  global sensitivity analysis methods and local sensitivity analysis methods, respectively.
%
%

As for global sensitivity analysis, since function evaluation of $f(x)$ is a very costly operation, we try to make the new design points are maximally informative. Since 
more experimental design points should usually be 
placed in regions
with finer detail and less in areas where the function is smoother, sequentialized designs or active learning are usually required. For example, Kriging based methods use its prediction errors to guide the arrangement of design points concentrated to the areas which need more exploration partially due to nonlinearity.  Most of the optimization algorithm,  considered as sequentialized designs, also have such features when choosing the function evaluation points.   However, one side effect of the sequentialized designs is often lack of enough global exploration in some complicated cases.

Correspondingly, it may also be desirable to locate some experimental design points in a way that
does not assumes any knowledge of the underlying function from the previous function evaluations, in order to more encourage global exploration.
In such cases,  one stage space filling methods can be used cover the whole domain evenly and avoid non-exploration of certain regions, for example, the widely used Latin Hypercube Sampling related methods. 
%
 A comprehensive survey about this topic was written by Shan et al. (2010)  \cite{Shan10SurveyHEB}.

 In this paper, we will not use a specific sequentialized strategy  to add extra design points, because 
 one main goal of this paper is to demonstrate that the available function evaluations during the optimization step play an important role in helping generate a satisfying response surface of global approximation.  Therefore we choose the widely used space filling experimental design methods such as  optimal Latin Hypercube (LHD, for short) \cite{Ye00OSLHD} which are model-independent, to extend the set of design points, though
 other design methods can be also adopted here.

As for local sensitivity analysis, when a surrogate of global high fidelity is available, it can be just performed on it. However, if $f(x)$ defined in a relatively high dimensional space with complicated input-output relationship, it is hard to get a surrogate of global approximation based on an affordable number of function evaluations. 
Correspondingly we proposed to 
develop a new tailored  design to reduce the size of the required experimental design points, by aiming for a surrogate of a local approximation which  may be only  suitable for the calculation of the targeted local sensitivity analysis quantities, but not fitful for other sensitivity methods.  In this paper, we take our new local sensitivity analysis method as an example and present a 
specific objective-oriented adaptive experimental design method for it.  The detailed description have already been presented in Section \ref{subsec:MVMLSDesign} and will be more details in Section  \ref{sec:benchmarkDesign}.



\subsubsection{Step 3: Establish a Surrogate Surface}
There are multiple response surface families: polynomials, splines, interpolating radial basis functions, kriging, generalized linear models, neural networks, regression trees, support vector machine, and many other nonparametric approaches, ect. We have not restricted our study to linear and quadratic functions because we know that there are strong interactions at multiple orders between input parameters in many scientific and engineering fields. Meanwhile we did not use the nonparametric methods because they are originally developed for situations with huge number of sample sizes whereas the main motivation of the usage of surrogates surfaces is to significantly reduce  the number of experimental design points.

For the relatively high dimensional ($n\ge 10$),  nonlinear, computationally-expensive black-box functions,  Kriging (Gaussian process regression) \cite{Cressie93Kriging,Sacks89DACE1}, Radial basis function (RBF) interpolation are mostly widely used for this kind of models. 
In addition, multiple kinds of surrogates can be chained together to approximate a large scale complex systems. In this paper, we will be using both the radial basis function interpolation and Kriging as the surrogate models \cite{Powell1992RBF} and compare their different performances in our cases. 


\subsubsection{Step 4: Perform Sensitivity Analysis on the Surrogate Surface}
 As for global sensitivity analysis, we will be using Extended FAST, which has been  introduced in Section \ref{subsec:EFAST}.  As for local sensitivity analysis, we test our new local sensitivity analysis method which has been  introduced in Section \ref{subsec:MVMLS}, and the formulas of corresponding sensitivity quantities are given in Section \ref{subsec:SVD}.  


\section{Numerical  Experiments}\label{Sec:NumExp}

\subsection{Introduction to Test problems}

Our method will be tested  on  $3$ typical examples, including two
synthetic problems and two from real applications, which  are all nonlinear problems and the number of variables is no less than $10$.  
%
The problem dimension  is much higher
than most of testing problems of the existing works. 

\paragraph{Test Problem $1$}
This  one originally appeared on the book of Hock and Schittkowski (1981) \cite{Hock80TestProblemsBook} to test nonlinear optimization algorithms.
It was also picked out by Jin and et al. \cite{Jin00comparativestudies} and Shan and Wang \cite{shan10HEB} to demonstrate the performance of their method to establish the surrogate model.
Unlike many other common synthetic testing problems used by the optimization community,  we make its independent variables  have drastically different sensitivities by setting much different  coefficients $c_i$  and therefore is suitable as a test problem for
sensitivity analysis. It is a  highly nonlinear problems with the following form
$$f(x)= \sum_{i=1}^{10} \exp(x_i)(c_i+x_i-\ln (\sum_{k=1}^{10}\exp(x_i)))$$ where
$c_{i=1,\ldots,10}=[ -35,  -28,  -20,  -16,  -10,  -6, -4,  -2,   -1,   -0.02]
$ in our testing problem, though other settings might also be acceptable; and $x_i \in [-1,1].$ 

\paragraph{Test Problem 2} This is a parameter calibration problem for the simulation of the Town Brook watershed which is a 37 km2 subregion of the Cannonsville watershed (1200 km2)
in the Catskill Region of New York State. The time series Y of measured stream flows
and total dissolved phosphorus (TDP) concentrations used in the analysis contains 1096
daily observations (from October 1997 to September 2000) based on readings by the U.S.
Geological Survey for water entering the West Branch of the Delaware River from the Town
Brook watershed. We used the SWAT2005 simulator (Arnold et al. (1998)), which has been
used by over a thousand agencies and academic institutions worldwide for the analysis of water flow and nutrient transport in watersheds (e.g., Eckhardt et al. (2002), Grizzetti et
al. (2003), Shoemaker et al. (2007), Tolson and Shoemaker (2007b)). The water draining the
Town Brook and rest of the Cannonsville watershed collects in the Cannonsville Reservoir,
from which it is piped hundreds of miles to New York City for drinking water. Water quality
is threatened by phosphorus pollution and, if not protected, could result in the need for a
New York City water filtration plant estimated to cost over $8$ billion. For this economic
reason as well as for general environmental concerns, there is great interest in quantifying
the parameter uncertainty for this model.

The input information of the Town Brook simulator is discussed briefly in Tolson and
Shoemaker (2007a) and in more details in Tolson and Shoemaker (2004, 2007b).


Here we only estimate $10$ flow related parameters, by minimizing the sum of squares errors between simulated flow data and
measured flow data.

$$\min f(x)=\sum_{i=1}^{2192} (Y_i-y_i(x))^2$$
where $x \in [0,1]^{10}$ is the parameter of the involved simulator $y(x)$; $Y$ and $y$ are the measured data  and the output of the simulator, respectively and each of it is a vector of length $2192$.

\paragraph{Test Problem 3}
It is a $36$-dimensional groundwater bioremediation application involving partial differential equations \cite{yoon99comparison}. Bioremediation is a process to remove organic compounds or to transform them to less harmful substances by utilizing the microorganism's catabolic (energy producing) and anabolic (cell synthesizing) activities. This process is enhanced by the injection of an electron acceptor (e.g., oxygen) or nutrients (e.g., phosphorous and nitrogen) to promote microbial growth.  Efficient in situ bioremediation design attempts to insure that the well locations and pumping rates are both economical and effective at distributing the electron acceptor or nutrients throughout the system.   For the objective function $f(x)$, the unknown variable $x \in \mathbb{R}^{n}$ represents the well locations and pumping rates and its domain has been normalized to $[0, 1]^n$, where $n=36$.

\subsection{Experimental Setup} \label{sec:benchmarkDesign}
In this section, we demonstrate the main points of this paper as follows: (1) function evaluations during the (global) optimization, an an adaptive experimental design,   are playing an important role to help generate a high quality surrogate of global approximation. (2) For  our new local sensitivity analysis, the novel corresponding tailored experimental design method is more efficient to generate a surrogate for calculation of its quantities than other state-of-art experimental design methods. (3) For relatively low dimensional problems, Kriging behaviors slightly worse than Gaussian RBF. For  relatively high dimensional problems, Kriging behaviors much worse than Gaussian RBF, when the number of experimental design points is small.  

%
%
%


\subsubsection{Global Sensitivity Analysis}
As for the first point, O3AED  uses the function evaluation points during the optimization step as the initial design points and then extends the
set of design points by other available experimental design methods, in order to obtain a surrogate model of 
good global approximation. Here spacial filling methods \cite{Jin2005268,Johnson90MiniMax,Tang93OALHD,Park94OptimalD,Morris95OptimalLHD,Ye98OCLHD,Palmer01MBLHD,Crary02MetaModel,Leary03OOALHD} such as the optimal Latin Hypercube Design (LHD, for short) \cite{Ye00OSLHD} is adopted to generate the extra design points in Step $2$, though other  design methods could be adopted.  
The number of function evaluations during optimization is denoted to be $N_{OPT}$.   The number of  the extra design points generated in Step $2$  is denoted as $N_{EXT}$.
  Then we generate a surrogate model $s(x)$ based on the $N_{OPT}+N_{EXT}$ function evaluations of $f(x)$ and  perform corresponding sensitivity analysis on it.   

One important feature of O3AED   is to reuse the function evaluations during the optimization step to help generate of a surrogate of good global approximation property.  Therefore, we are comparing O3AED  with other classic experimental designs for generating a well-approximating surrogate, without the adoption of optimization. Here 
we use the optimal LHD \cite{Ye00OSLHD} and it  generates $N_{OPT}+N_{EXT}$ design points for fair comparison.  The comparison
of O3AED with LHD in terms of Extended FAST is performed in Test Problems $1$ and $2$ in Section \ref{subsec:NumResults}. The important role of function evaluations during optimization which can be considered as an adaptive experimental design,  is well demonstrated.  In addition, in Test Problem $2$, we also compare the performance of the Gaussian RBF and Kriging as surrogate surfaces.



Finally,  as for Extended FAST, besides the above two surrogate based methods, another alternative is to directly calculate its quantities  using $N_{OPT}+N_{EXT}$  samples on $f(x)$, without the help of surrogates. This method is denoted as ``DIRECT''.
As we have known, Extended FAST typically requires a huge number of function evaluations in order for the accuracy of the calculated sensitivity quantities, and such an alternative is expected to have a poor accuracy. 


\subsubsection{Local Sensitivity Analysis}


 As mentioned before,  for local sensitivity analysis, it is often wasteful to get a surrogate of a good global approximation property, or even impossible to establish such a surrogate due to a very limited number of function evaluations of $f(x)$, especially for relatively high dimensional problems with complex input-output relationships.  Therefore 
we use  the specific objective-oriented adaptive 
design introduce in Section \ref{subsec:MVMLS} in Step $2$ of O3AED  to add these extra design points. 
Specifically, when we extended design points are uniformly randomly picked a
small portion of  MVMSL points.  Instead of focusing on the global approximation of the surrogate over the whole domain,  our new experimental design method tries to establish a surrogate, which is only of good approximation at the MVMSL points. So it is expected to require less design points than LHD when the same accuracy at the MVMSL points are achieved. 
 The advantages of O3AED in terms of this kind of ``objective oriented" experimental design over the optimal LHD to add these extra design points are illustrated via Test Problem  $3$ in Section \ref{subsec:NumResults}. 

\subsection{Evaluation criteria}
We  evaluate the their performances of different experimental design methods and surrogate surfaces, by comparing their calculated sensitivity quantities with the true or ``gold standard'' sensitivity quantities.
 For Extended FAST, the analytic values of  $ST_i$ and $S_i$ are often hard or impossible to obtain, especially for black-box functions. Therefore, in order to compare the performances of the above algorithms, we execute a large number (for example, $10000\times d$) of evaluations of $f(x)$ in order to obtain good estimate of them and take these estimates as the ``gold standard'' or references, though  such a large number of evaluations are often computationally prohibitive for  practical computationally expensive functions. The second column of Table \ref{Tab: N_FunEvals} is the number of  the performed function evaluations of $f(x)$ when calculating the ``gold standard'' Extended FAST quantities.  As for MVMLS, their true  values are those calculated on $f(x)$, instead of the surrogate $s(x)$. The sixth column of Table \ref{Tab: N_FunEvals} is the number of  the performed function evaluations of $f(x)$ when calculating the true MVMLS quantities for one given perturbation step.




Correspondingly, we evaluate different computational methods by calculating  Relative Error (Rel\_Err) of the calculated sensitivity measures.
\begin{equation}Rel\_Err (M)=\frac{\sqrt{\sum_{i}^n (S_i^{M}-S_i^{R})^2}}{\sqrt{\sum_{i}^n (S_i^{R})^2}}, \end{equation}
where $S^{M}$ is  sensitivity quantity calculated by either surrogated assisted methods based on O3AED and LHD, or ``DIRECT'' method, and  $S_i^{R}$ is the ``gold standard'' value of the corresponding sensitivity quantity.  The smaller Rel\_Err, the better the corresponding computational method is. The third, fourth and fifth columns of Table \ref{Tab: N_FunEvals} are number of function evaluations of $f(x)$ performed by O3AED , LHD and ``DIRECT'' for the Extended FAST, respectively.  The seventh and eighth columns of Table \ref{Tab: N_FunEvals} are number of function evaluations of $f(x)$ performed by O3AED, and LHD for MVMSL (for one perturbation step), respectively. 

As for MVMSL, given a perturbation step $\rho$, each parameter might have $3$ sensitivity indices $SI_{i}^{1, \rho}, SI_{i}^{2, \rho}$, and $SI_{i}^{3, \rho}$, based on one-at-a-time perturbation, two-at-a-time perturbation, and three-at-a-time perturbation, respectively.  That is to say, $S_i^{M}$ and $S_i^{R}$ can be calculated values of  $SI_{i}^{1, \rho}$,  or $SI_{i}^{2, \rho}$, or $SI_{i}^{3, \rho}$, calculated based on the true function $f(x)$ and the surrogate $s(x)$, respectively.

As for MVMSL, we also would like to introduce another evaluation criteria, called matching rate. Basically, we evaluate whether the $s_3$ ($s_3=100$, for example) most sensitive duos or triples (for example, $SI_{(k^{+},j^+, i^+)}^{3,\rho}$) can be detected based on the surrogate, since we are usually interested in them. The matching rate is the measurement or  the percentage of the correctness.
%
Specifically, for each perturbation $\rho$, since we had calculated $SI_{(k^{+},j^+, i^+)}^{3,\rho},$ $SI_{(k^{+},j^-, i^+)}^{3,\rho}$ and etc,  we can
sorted them for most sensitive to least sensitive. We compare the first $s_3$ ($s_3=100$, for example) most sensitive one calculated by the response surface $s(x)$ with those
based on the true function $f(x)$, and calculate its matching rate, denoted as $\gamma_3^{\rho}$, which is the number of correctly detected over $s_3$ (i.e. $0\le \gamma_3^{\rho} \le 1$).  The higher $\gamma_2^{\rho}$ or $\gamma_3^{\rho}$, the better quality of the generated surrogate $s(x)$.


%

\begin{table}[htbp]\caption{Table of comparison of numbers of expensive function evaluations required for different experimental design methods}\label{Tab: N_FunEvals}
\centering
\scriptsize{
\begin{tabular}{|l|l|l|l|l|l|l|l|l|l|l|l|l|l|l|l|l|l|l|l|l|}
  \hline
   \multirow{2}{*}{{\tiny Problems}}& \multicolumn{4}{|c|}{EFAST} &  \multicolumn{3}{|c|}{MVMSL}
  \\\cline{2-8}  &Ref & O3AED  &  LHD & DIRECT& Ref & O3AED   &LHD
  \\\hline
1 &100,000 &100 & 100 & 650 &$\backslash$ & $\backslash$&$\backslash$\\\hline
2 &100,000 & 650& 650 &650  & $\backslash$ & $\backslash$&$\backslash$ \\\hline
3 &$\backslash$ &$\backslash$&$\backslash$ & $\backslash$ & 59,712& 3732& 3732\\
\hline
\end{tabular}}
\end{table}

\subsection{Experimental Results} \label{subsec:NumResults}

\subsubsection{Test Problem 1} \label{subsec:examaple1}
The optimization algorithm applied to this test problem is the multistart pattern search method. The particular multistart approach we used was multi level single linkage (MLSL) method  \cite{Rinnooy97MLSL} and the pattern search algorithm \cite{Torczon97Pattern} was implemented in the Matlab Genetic Algorithm and Direct Search Toolbox.   The maximum number $N_{OPT}$ of function evaluation of $f(x)$ is $100$.
For the Extended FAST, O3AED only uses the function evaluations during the optimization to generate the surrogate model and no extra design points are generated, which means that $N_{EXT}=0.$, i.e. Step $2$ of O3AED  is skipped.  For fair comparing
the number of design points of the optimal LHD is therefore $N_{OPT}+N_{EXT}=100$. For ``DIRECT" method,  we are using $650$ samples on $f(x)$, instead of $N_{OPT}+N_{EXT}=100$, because the minimum number of samples for the Extended FAST implemented by Facilia \cite{Ekstrom05Eikos} is $65n=650$, where $n$ is the dimension of the problem and $n=10$ here.

The results of the Extended FAST are showed in Table \ref{Tab: EFAST_Jin}.
For each method, we ranked the
 sensitivity quantities in a descending order and listed the corresponding input variables.  We also calculated the RSSE (Relative Root of Sum of Square Errors) of the calculated sensitivity quantities. The smaller RSSE, the better the method is expected to be.  We can see that our method O3AED  is the best among the three computationally feasible candidate methods. Notice that even though $650$ samples are used for the ``DIRECT" method  rather than $100$ samples, its performance is still much worse than O3AED, partially because it fails to take advantages of the underlying smoothness of $f(x)$, which is well made use of by the establishment of a surrogate.

\begin{table}[htbp]\caption{Extended FAST result on the Test Problem $1$: ``Ref'' stands for the method to calculate the reference sensitivity values using $10000 n=100000$ true function evaluations of $f(x)$.   
$N_{OPT}=100$; $N_{EXT}=0.$ ``DIRECT" uses $650$ true function evaluations.}\label{Tab: EFAST_Jin}
\centering
\scriptsize{
\begin{tabular}{|l|l|l|l|l|l|l|l|l|l|l|l|l|l|l|l|l|l|l|l|l|}
  \hline
   \multirow{2}{*}{{\tiny Rank}}& \multicolumn{2}{|c|}{Ref} &  \multicolumn{2}{|c|}{O3AED} & \multicolumn{2}{|c|}{LHD}& \multicolumn{2}{|c|}{DIRECT}& \multicolumn{2}{|c|}{Ref}& \multicolumn{2}{|c|}{O3AED}& \multicolumn{2}{|c|}{LHD}&
   \multicolumn{2}{|c|}{DIRECT}
  \\\cline{2-17}  & {i} &{\tiny $ST_i$} &  {i} &{\tiny $ST_i$}& {i} &{\tiny $ST_i$}& {i} &{\tiny $ST_{i}$}& {i} &{\tiny $S_{i}$}&{i} &{\tiny
 $S_{i}$}&
  {i} &{\tiny $S_{i}$}& {i} &{\tiny $S_{i}$}\\\hline
1 &1 &0.405 & 1 & 0.403 & 9& 0.229 & 1 & 0.630 & 1 & 0.403  & 1 & 0.402 & 4 & 0.022& 1 & 0.623\\
2 &2 &0.264 & 2 & 0.263 & 8& 0.214 & 2 & 0.402 & 2 & 0.263  & 2 & 0.261 & 3 & 0.018& 2 & 0.395\\
3 &3 &0.145 & 3 & 0.146 & 10& 0.210 & 3 & 0.213 & 3 & 0.145  & 3 & 0.146 & 10 & 0.018& 3 & 0.208\\
4 &4 &0.101 & 4 & 0.103 & 7& 0.210 & 4 & 0.147 & 4 & 0.100  & 4 & 0.102 & 1 & 0.018& 4 & 0.141\\
5 &5 &0.045 & 5 & 0.046 & 4& 0.208 & 6 & 0.079 & 5 & 0.045  & 5 & 0.046 & 8 & 0.014& 5 & 0.068\\
6 &6 &0.019 & 6 & 0.019 & 3& 0.197 & 5 & 0.074 & 6 & 0.019  & 6 & 0.019 & 5 & 0.011& 6 & 0.066\\
7 &7 &0.011 & 7 & 0.012 & 1& 0.195 & 7 & 0.029 & 7 & 0.011  & 7 & 0.011 & 9 & 0.009& 7 & 0.023\\
8 &8 &0.006 & 8 & 0.006 & 5& 0.179 & 9 & 0.024 & 8 & 0.006  & 8 & 0.005 & 6 & 0.009& 9 & 0.009\\
9 &9 &0.003 & 9 & 0.004 & 6& 0.170 & 10 & 0.018 & 9 & 0.003  & 9 & 0.003 & 7 & 0.008& 8 & 0.008\\
10 &10 &0.002 & 10 & 0.002 & 2& 0.144 & 8 & 0.017 & 10 & 0.002  & 10 & 0.002 & 2 & 0.008& 10 & 0.005\\\hline
{\tiny Rel\_Err} & &  & & 0.003 & & 0.539 &  & 0.286 &  &   &  & 0.003 &  & 0.863&  & 0.274
\\
\hline
\end{tabular}}
\end{table}

\subsubsection{Test Problem $2$}
The optimization algorithm applied to this test problem is  the stochastic RBF optimization algorithm, with the maximum number $N_{OPT}$ of function evaluation of $f(x)$ being $150$. Extended FAST is performed to compare O3AED  and two other alternatives.


            O3AED   adds $500$ extra design points  by the optimal LHD in Step 2, i.e. $N_{EXT}=500.$  Correspondingly, for the alternative, the pure optimal LHD, the number of design points is therefore $N_{OPT}+N_{EXT}=650$ for fair comparison and we use the Gaussian RBF as the surrogate surface. For the ``DIRECT" Method,  we are still using $650$ samples on $f(x)$. For our experimental design method, we compare the performance of using Gaussian RBF and Kriging as the surrogate surfaces.
            The results are showed in Tables \ref{Tab: EFAST_TB2} and \ref{Tab: EFAST_TB2-2}, where the $ST_i$ and $S_i$ are presented, respectively.
For each method, we sorted the calculated values of the sensitivity quantities from largest to smallest and the corresponding parameter index. 
We also calculated the RSSE (relative root of square errors) of the calculated sensitivity quantities.
 The smaller Rel\_Err, the better the method is.  We can see that our experimental design method O3AED is the best among the three candidate methods.  This experiment verify that the function evaluations of optimization helps capture the global shape of  $f(x)$ in an effective way. In addition, we also observed that the global approximation performance of Kriging is not as good as Gaussian RBF using the same experimental design in this case. 




\begin{table}[htbp] \caption{Extended FAST result on the Test Problem $2$: ``Ref'' stands for the method to calculate the reference sensitivity values using $10000 n=100000$ true function evaluations of $f(x)$.  
``Val'' and ``Rank'' stand for the calculated values of sensitivity indices and the corresponding parameter ranks using different methods. $N_{OPT}=150$; $N_{EXT}=500$. 
}\label{Tab: EFAST_TB2}
\centering
\scriptsize{
\begin{tabular}{|l|l|l|l|l|l|l|l|l|l|l|l|l|l|l|l|l|l|l|l|l|}
 \hline
   \multirow{3}{*}{{\tiny Rank}}& \multicolumn{10}{|c|}{$ST_i$}  \\\cline{2-11}
   & \multicolumn{2}{|c|}{Ref} &  \multicolumn{2}{|c|}{O3AED\_RBF} & \multicolumn{2}{|c|}{O3AED\_Kriging} &\multicolumn{2}{|c|}{LHD\_RBF}& \multicolumn{2}{|c|}{DIRECT}
  \\\cline{2-11}  & {i} &{\tiny Val(i)} &  {i} &{\tiny Val(i)}& {i} &{\tiny Val(i)}& {i} &{\tiny Val(i)}& {i} &{\tiny Val(i)} \\\hline
1 &9 &0.627 & 9 & 0.610 & 9&0.616 & 9& 0.528 & 9 & 0.546 \\
2 &5 &0.266 & 5 & 0.286 & 5& 0.328& 5& 0.242 & 5 & 0.305 \\
3 &10 &0.064 & 10 & 0.061 &10 &0.060 & 10& 0.065 & 3 & 0.103 \\
4 &2 &0.060 & 2 & 0.054 &2 &0.050 & 2& 0.044 & 10 & 0.100  \\
5 &3 &0.046 & 1 & 0.037 &1 & 0.039& 1& 0.036 & 1 & 0.087 \\
6 &1 &0.040 & 3 & 0.030 &3 &0.033 & 3& 0.028 & 2 & 0.071 \\
7 &4 &0.033 & 4 & 0.015 & 4& 0.018& 4& 0.016 & 4 & 0.053 \\
8 &6 &0.007 & 6 & 0.009 &6 &0.009 & 6& 0.012 & 7 & 0.047 \\
9 &8 &0.004 & 7 & 0.007 &7 &0.007 & 7& 0.008 & 8 & 0.046 \\
10 &7 &0.004 & 8 & 0.005 & 8&0.006 & 8& 0.006 & 6 & 0.041 \\\hline
{\tiny Rel\_Err} & &  & & 0.032 & &0.058 & & 0.093 &  & 0.123 \\
\hline
\end{tabular}}
\end{table}

\begin{table}[htbp] \caption{Extended FAST result on the Test Problem $2$: ``Ref'' stands for the method to calculate the reference sensitivity values using $10000 n=100000$ true function evaluations of $f(x)$.  
``Val'' and ``Rank'' stand for the calculated values of sensitivity indices and the corresponding parameter ranks using different methods. $N_{OPT}=150$; $N_{EXT}=500$. 
}\label{Tab: EFAST_TB2-2}
\centering
\scriptsize{
\begin{tabular}{|l|l|l|l|l|l|l|l|l|l|l|l|l|l|l|l|l|l|l|l|l|}
 \hline
   \multirow{3}{*}{{\tiny Rank}}&  \multicolumn{10}{|c|}{$S_i$} \\\cline{2-11}
   & \multicolumn{2}{|c|}{Ref} &  \multicolumn{2}{|c|}{O3AED\_RBF} & \multicolumn{2}{|c|}{O3AED\_Kriging} & \multicolumn{2}{|c|}{LHD\_RBF}& \multicolumn{2}{|c|}{DIRECT}
 \\\cline{2-11}  & {i} &{\tiny Val(i)} &  {i} &{\tiny Val(i)}& {i} &{\tiny Val(i)}& {i} &{\tiny Val(i)}& {i} &{\tiny Val(i)} \\\hline
1  & 9 & 0.530  & 9 & 0.523 & 9&0.533 & 9 & 0.464& 9 & 0.449\\
2  & 5 & 0.215  & 5 & 0.222 & 5&0.249 & 5 & 0.192& 5 & 0.207\\
3  & 10 & 0.034  & 10 & 0.038 & 10& 0.037& 10 & 0.036& 10 & 0.042\\
4  & 3 & 0.020  & 3 & 0.020 &3 &0.020 & 3 & 0.017& 3 & 0.038\\
5  & 1 & 0.013  & 1 & 0.015 &1 &0.015 & 1 & 0.015& 1 & 0.024\\
6 & 2 & 0.009  & 2 & 0.011 & 2& 0.012& 2 & 0.008& 2 & 0.013\\
7  & 4 & 0.005  & 4 & 0.003 &4 &0.003 & 6 & 0.004& 6 & 0.006\\
8  & 6 & 0.002  & 6 & 0.003 &6 &0.003 & 4 & 0.004& 4 & 0.005\\
9  & 8 & 0.000  & 7 & 0.001 &7 &0.001 & 7 & 0.001& 8 & 0.004\\
10 & 7 & 0.000  & 8 & 0.001 & 8&0.000 & 8 & 0.000& 7 & 0.003\\\hline
{\tiny Rel\_Err}  &  &   &  & 0.014 & &0.041 &  & 0.107&  & 0.103\\
\hline
\end{tabular}}
\end{table}

\subsubsection{Test Problem $3$}
This test problems has $36$ parameters and is used to test the performance of O3AED  in terms of calculating MVMSL quantities. In this example, we show the advantages of using ``objective oriented" adaptive experimental design to add extra experiment design points of Step $2$ of O3AED. Specifically, in this example, we adopted the local stochastic RBF method \cite{Regis2007SRBF} where $N_{OPT}=600$ function evaluations were executed and set $\rho=0.2$ as an example, though other value of $\rho$ might be also applicable. In Step $2$ of  O3AED, we set all the $2n$ MVMSL univariate perturbation points, and $45n$ randomly picked MVMSL two-variable perturbation points and $40n$ randomly picked MVMSL three-variable perturbation points as the experimental design points to establish a surrogate. Therefore, O3AED  adds $2n+45n+40n=3132$ function evaluations of $f(x)$. Notice that without surrogate, $59712$ function evaluations of $f(x)$ is required, which is around $16$ times computational cost than the above two methods based on the surrogate models.

 As an alternative, ones can add the extra experimental design points using the optimal LHD, without considering the definition of MVMSL. For fair comparison,  the same number ($2n+45n+40n$=3131) design points through the optimal LHD  within the neighborhood ( $\rho$=0.2) of the optimal solution are generated.

 Tables \ref{Tab: MVMLS_GWB1} and  \ref{Tab: MVMLS_GWB1-2} show the obtained MVMLS quantities  based on the surrogate surfaces based on  O3AED and LHD, which are the two ways to generate the extra experimental design points besides those of the optimization procedure. 
In addition, we also compare the performance of Gaussian RBF and Kriging for the experimental design method ``O3AED". 

We see that the results by O3AED and Gaussian RBF  are much more accurate than that of the combination of  O3AED and Kriging. It shows that in cases of high dimensional nonlinear problems and very few experimental design points, the performance of Kriging degrades a lot. Furthermore, we can see that 
  the MVMSL-specific design adopted in O3AED to add the extra design points, behaviors much better than the optimal LHD. 

 Table \ref{Tab: MVMLS_GWB2} shows the obtained sensitivity quantity of  $x_i$ based on the absolute value  of the $i$th component of the eigenvector of the matrix $H^{\rho}$ using $\rho=0.2$. Here we consider the eigenvectors $U^{1,\rho}$ and $U^{2,\rho}$, which correspond to the two eigenvalues of the largest magnitude and second largest magnitude, respectively. We also calculated the Rel\_Err (relative root of square errors) of the calculation of  these two eigenvectors.
 We can see that O3AED is much better than LHD, because its resulted rank is more close to the true rank. In addition, the calculated sensitivity quantities corresponding to  O3AED have a much smaller Rel\_Err.  Therefore, we can see that the sensitivity-specific design ``O3AED" is more promising. 

\begin{table}[htbp]\caption{Comparison of the MVMSL sensitivity quantities calculated by O3AED  and LHD;  $\rho=20\%$;  ``Ref'' represents the true sensitivity values calculated on $f(x)$.}\label{Tab: MVMLS_GWB1}
\centering
\scriptsize{
\begin{tabular}{|l|l|l|l|l|l|l|l|l|l|l|l|l|l|l|l|l|l|l|l|l|l|l|l|l|l|ll|}
  \hline
    \multirow{2}{*}{{\tiny  Rank}} & \multicolumn{2}{|c|}{Ref} &  \multicolumn{2}{|c|}{O3AED\_RBF} & \multicolumn{2}{|c|}{LHD\_RBF}& \multicolumn{2}{|c|}{O3AED\_Kriging}\\\cline{2-9}
   &i & \tiny{$SI_{i}^{1, \rho}$} &i & \tiny{$SI_{i}^{1, \rho}$} & i &\tiny{$SI_{i}^{1, \rho}$} & i & \tiny{$SI_{i}^{1, \rho}$} \\\hline
1& 8&  0.69& 8 & 0.69&7 & 0.69& 8 &  0.51 \\
2& 7&  0.56& 7 & 0.56&8 & 0.61& 7 &  0.40 \\
3& 21&  0.51& 21 & 0.51&21 & 0.55& 24 &  0.35 \\\hline
4& 24&  0.48& 24 & 0.48&15 & 0.49& 21 &  0.35 \\
5& 18&  0.48& 18 & 0.48&18 & 0.42& 18 &  0.32 \\
6& 27&  0.43& 27 & 0.43&36 & 0.41& 27 &  0.28 \\
7& 15&  0.40& 15 & 0.40&24 & 0.38& 15 &  0.26 \\
8& 9&  0.36& 9 & 0.36&27 & 0.38& 9 &  0.22 \\\hline
9& 12&  0.28& 12 & 0.28&33 & 0.36& 36 &  0.18 \\
10& 30&  0.26& 30 & 0.26&20 & 0.35& 12 &  0.14 \\
11& 36&  0.26& 36 & 0.26&9 & 0.34& 14 &  0.13 \\
12& 11&  0.24& 11 & 0.24&5 & 0.34& 11 &  0.13 \\
13& 23&  0.23& 23 & 0.23&17 & 0.32& 30 &  0.13 \\
14& 14&  0.23& 14 & 0.23&13 & 0.30& 20 &  0.12 \\
15& 20&  0.21& 20 & 0.21&30 & 0.29& 23 &  0.11 \\\hline
16& 17&  0.16& 17 & 0.16&35 & 0.22& 34 &  0.09 \\
17& 6&  0.15& 6 & 0.15&14 & 0.21& 4 &  0.09\\
18& 33&  0.12& 33 & 0.12&32 & 0.19& 31 &  0.09 \\
19& 35&  0.10& 35 & 0.10&12 & 0.19& 28 &  0.09 \\
20& 3&  0.09& 3 & 0.09&4 & 0.18& 33 &  0.09 \\
21& 29&  0.09& 29 & 0.09&23 & 0.17& 17 &  0.09 \\
22& 13&  0.08& 13 & 0.08&22 & 0.13& 10 &  0.08 \\
23& 16&  0.08& 16 & 0.08&26 & 0.12& 1 &  0.08 \\
24& 26&  0.07& 26 & 0.07&6 & 0.12& 2 &  0.08 \\
25& 22&  0.04& 22 & 0.04&3 & 0.11& 35 &  0.08 \\\hline
26& 32&  0.03& 32 & 0.03&16 & 0.10& 25 &  0.08 \\
27& 19&  0.03& 19 & 0.03&19 & 0.09& 5 &  0.08 \\
28& 5&  0.03& 5 & 0.03&25 & 0.08& 19 &  0.08 \\
29& 25&  0.03& 25 & 0.03&28 & 0.08& 32 &  0.08 \\
30& 2&  0.02& 2 & 0.02&1 & 0.07& 22 &  0.07 \\
31& 1&  0.02& 1 & 0.02&11 & 0.07& 6 &  0.06 \\
32& 10&  0.02& 10 & 0.02&29 & 0.07& 26 &  0.04 \\
33& 28&  0.02& 28 & 0.02&31 & 0.07& 16 &  0.04 \\
34& 31&  0.02& 31 & 0.02&2 & 0.05& 13 &  0.04\\
35& 4&  0.01& 4 & 0.01&10 & 0.04& 3 &  0.03 \\
36& 34&  0.01& 34 & 0.01&34 & 0.03& 29 &  0.03 \\\hline
{\tiny Rel\_Err}  & &  & & 0.000 & & 0.229 &  & 0.343  \\

\hline
\end{tabular}}
\end{table}

\begin{table}[htbp]\caption{Comparison of the MVMSL sensitivity quantities calculated by O3AED  and LHD;  $\rho=20\%$;  ``Ref'' represents the true sensitivity values calculated on $f(x)$.}\label{Tab: MVMLS_GWB1-2}
\centering
\scriptsize{
\begin{tabular}{|l|l|l|l|l|l|l|l|l|l|l|l|l|l|l|l|l|l|l|l|l|l|l|l|l|l|ll|}
  \hline
    \multirow{2}{*}{{\tiny  Rank}} &   \multicolumn{2}{|c|}{Ref}& \multicolumn{2}{|c|}{O3AED\_RBF}& \multicolumn{2}{|c|}{LHD\_RBF}& \multicolumn{2}{|c|}{O3AED\_Kriging} & \multicolumn{2}{|c|}{Ref}&
   \multicolumn{2}{|c|}{O3AED\_RBF}&\multicolumn{2}{|c|}{LHD\_RBF}  &\multicolumn{2}{|c|}{O3AED\_Kriging}\\\cline{2-17}
   &i  & \tiny{$SI_{i}^{2, \rho}$} & i& \tiny{$SI_{i}^{2, \rho}$} & i &\tiny{$SI_{i}^{2, \rho}$}&i& \tiny{$SI_{i}^{2, \rho}$} & i& \tiny{$SI_{i}^{3, \rho}$} &i & \tiny{$SI_{i}^{3, \rho}$} & i&\tiny{$SI_{i}^{3, \rho}$} & i&\tiny{$SI_{i}^{3, \rho}$} \\\hline
1&  8 &  0.86 &8 &  0.85& 7&  0.86 & 8& 0.62& 8 & 1.04& 8&  1.06 & 7 & 1.03& 15&0.84\\
2&  7 &  0.75 &7 &  0.75& 8&  0.79 &7 &0.56 & 7 & 0.93& 7&  0.96 & 8 & 0.96& 7&0.68\\
3&  21 & 0.70 &21 &  0.69& 21&  0.74 &15 & 0.56& 21 & 0.88& 21&  0.90 & 21 & 0.92&1 &0.68\\\hline
4&  18 & 0.66 &24 &  0.66& 15&  0.66 & 21&0.53 &18 & 0.85& 24&  0.88 & 15 & 0.85& 8&0.67\\
5&  24 & 0.66 &18 &  0.66& 18&  0.62 &18 &0.50 &24 & 0.84& 18&  0.88 & 18 & 0.81& 3&0.66\\
6&  27 & 0.63 &27 &  0.63& 36&  0.61 &24 &0.50 &27 & 0.82& 27&  0.86 & 27 & 0.81& 29&0.65\\
7&  15 & 0.59 &15 &  0.59& 27&  0.61 & 27&0.46 &15 & 0.77& 15&  0.80 & 36 & 0.80& 9&0.60\\
8&  9 &  0.55 &9 &  0.54& 24&  0.60 &9 &0.44 &9 & 0.74& 9&  0.76 & 24 & 0.79& 18&0.60\\\hline
9&  30 &  0.47 &12 &  0.47& 33&  0.55 & 36& 0.35&30 & 0.67& 30&  0.69 & 9 & 0.73&28&0.59\\
10& 36 &  0.47 &30 &  0.46& 9&  0.54 &20 &0.34 &36 & 0.66& 36&  0.67 & 33 & 0.73&20&0.59\\
11&  12 &  0.46 &36 &  0.46& 5&  0.52 &3 &0.34 &12 & 0.65& 12&  0.67 & 17 & 0.70&21&0.57\\
12&  23 &  0.43 &11 &  0.42& 17&  0.51 & 12& 0.33&23 & 0.61& 23&  0.64 & 30 & 0.70&33&0.56\\
13&  11 &  0.41 &23 &  0.42& 20&  0.50 & 29& 0.33&11 & 0.58& 11&  0.63 & 5 & 0.69&24&0.54\\
14&  14 &  0.40 &20 &  0.41& 30&  0.50 &1 &0.32 &14 & 0.58& 20&  0.62 & 20 & 0.68&17&0.52\\
15&  20 &  0.40 &14 &  0.41& 13&  0.47 &33 & 0.32&20 & 0.58& 14&  0.62 & 14 & 0.67&27&0.51\\\hline
16& 17 &  0.35 &17 &  0.35& 14&  0.46 &30 & 0.32&17 & 0.54& 17&  0.57 & 35 & 0.66&14&0.51\\
17&  6 &  0.34 &6 &  0.34& 35&  0.45 &11 & 0.32&6 & 0.53& 6&  0.53 & 13 & 0.65&11&0.50\\
18&  33 &  0.31 &33 &  0.32& 12&  0.40 & 23& 0.31&3 & 0.50& 3&  0.52 & 12 & 0.61&12&0.50\\
19&  3 &  0.30 &3 &  0.31& 23&  0.39 &14 &0.30 &33 & 0.50& 33&  0.52 & 23 & 0.60&36&0.50\\
20&  29 &  0.28 &35 &  0.29& 6&  0.37 &28 &0.28 &35 & 0.47& 13&  0.49 & 6 & 0.58&23&0.49\\
21&  35 &  0.28 &13 &  0.28& 16&  0.36 &17 &0.27 &29 & 0.47& 35&  0.49 & 16 & 0.58&19&0.48\\
22&  16 &  0.28 &29 &  0.28& 26&  0.35 &6 &0.26 &16 & 0.46& 29&  0.49 & 22 & 0.55&30&0.47\\
23&  13 &  0.28 &16 &  0.28& 22&  0.35 & 16& 0.24&13 & 0.46& 16&  0.48 & 26 & 0.55&6&0.46\\
24& 26 &  0.27 &26 &  0.26& 11&  0.32 &26 &0.24 &26 & 0.46& 26&  0.46 & 11 & 0.54&16&0.46\\
25&  22 &  0.23 &22 &  0.23& 25&  0.32 & 13& 0.23&22 & 0.41& 22&  0.43 & 25 & 0.52&13&0.46\\\hline
26&  25 &  0.21 &32 &  0.21& 32&  0.31 & 22& 0.22&25 & 0.39& 32&  0.40 & 29 & 0.51&26&0.46\\
27&  32 &  0.21 &25 &  0.21& 29&  0.30 &31 &0.21 &32 & 0.39& 2&  0.39 & 3 & 0.49&2&0.45\\
28& 19 &  0.21 &19 &  0.21& 3&  0.30 &35 & 0.21&19 & 0.38& 25&  0.39 & 2 & 0.49&31&0.45\\
29&  2 &  0.20 &5 &  0.20& 2&  0.29 &19 &0.21 &2 & 0.38& 5&  0.39 & 32 & 0.49&22&0.44\\
30&  5 &  0.20 &2 &  0.20& 10&  0.29 &10 &0.20 &34 & 0.38& 19&  0.39 & 10 & 0.49&10&0.43\\
31&  1 &  0.20 &1 &  0.20& 34&  0.26 &2 &0.19 &5 & 0.37& 1&  0.38 & 34 & 0.46&5&0.40\\
32&  10 &  0.20 &10 &  0.20& 4&  0.26 & 25& 0.19&1 & 0.37& 28&  0.38 & 28 & 0.45&32&0.39\\
33&  34 &  0.19 &28 &  0.20& 28&  0.26 &5 & 0.18&31 & 0.37& 10&  0.38 & 19 & 0.44&35&0.39\\
34&  31 &  0.19 &31 &  0.20& 19&  0.25 & 32& 0.17&10 & 0.37& 34&  0.38 & 31 & 0.43&25&0.39\\
35& 28 &  0.19 &34 &  0.19& 1&  0.24 & 4& 0.16&28 & 0.37& 31&  0.38 & 1 & 0.43&34&0.37\\
36&  4 &  0.18 &4 &  0.19& 31&  0.24 & 34& 0.15&4 & 0.36& 4&  0.37 & 4 & 0.42&4&0.36\\\hline
{\tiny Rel\_Err}   &  &  &  & 0.014  &  & 0.174 & &0.212 &  & &  & 0.041 & & 0.144 & &0.166\\
\hline
\end{tabular}}
\end{table}

\begin{table}[htbp]\caption{The sorted eigenvector components of $H^{\rho}$ in descend order in terms of absolute values.  From column 2 to column 7 is related with the eigenvector corresponding to the eigenvalue of the largest absolute value.   From column 8 to column 13 is related with the eigenvector corresponding to the eigenvalue of the second largest absolute value.
$\rho=20\%$;  ``Ref'' represents that the Hessian matrix is directly calculated on $f(x)$. }\label{Tab: MVMLS_GWB2}
\centering
\scriptsize{
\begin{tabular}{|l|l|l|l|l|l|l|l|l|l|l|l|l|l|l|l|l|l|l|l|l|l|l|l|l|l|ll|}
  \hline
    \multirow{2}{*}{{\tiny  Rank}} & \multicolumn{2}{|c|}{Ref} &  \multicolumn{2}{|c|}{O3AED} & \multicolumn{2}{|c|}{LHD}& \multicolumn{2}{|c|}{Ref}& \multicolumn{2}{|c|}{O3AED }& \multicolumn{2}{|c|}{LHD} \\\cline{2-13}
   &i & \tiny{$SI_{i}^{E, 1,\rho}$} &i & \tiny{$SI_{i}^{E, 1,\rho}$} & i &\tiny{$SI_{i}^{E, 1,\rho}$}& i & \tiny{$SI_{i}^{E, 2,\rho}$} & i& \tiny{$SI_{i}^{E, 2,\rho}$} & i &\tiny{$SI_{i}^{E, 2,\rho}$}\\\hline
1& 8&  0.34& 8 & 0.34&8 & 0.27& 8&  0.56& 8 & 0.55&36 & 0.34\\
2& 24&  0.28& 24 & 0.29&21 & 0.26& 7&  0.43& 7 & 0.42&33 & 0.33\\
3& 21&  0.27& 21 & 0.28&7 & 0.25& 13&  0.18& 29 & 0.19&5 & 0.31\\
4& 7&  0.27& 7 & 0.27&36 & 0.24& 26&  0.18& 23 & 0.17&21 & 0.28\\
5& 27&  0.26& 27 & 0.26&27 & 0.22& 29&  0.18& 16 & 0.17&14 & 0.24\\
6& 18&  0.25& 18 & 0.25&24 & 0.21& 16&  0.17& 13 & 0.17&24 & 0.23\\
7& 9&  0.24& 9 & 0.23&15 & 0.21& 23&  0.15& 26 & 0.16&35 & 0.22\\
8& 15&  0.23& 15 & 0.23&18 & 0.21& 22&  0.15& 22 & 0.15&7 & 0.22\\
9& 23&  0.19& 23 & 0.19&20 & 0.20& 17&  0.15& 19 & 0.15&20 & 0.22\\
10& 30&  0.19& 30 & 0.19&33 & 0.20& 10&  0.15& 10 & 0.15&6 & 0.22\\
11& 14&  0.19& 11 & 0.19&9 & 0.20& 25&  0.15& 20 & 0.15&27 & 0.21\\
12& 11&  0.19& 14 & 0.19&5 & 0.19& 19&  0.14& 25 & 0.15&3 & 0.19\\
13& 20&  0.19& 20 & 0.19&14 & 0.19& 32&  0.14& 32 & 0.14&2 & 0.19\\
14& 12&  0.18& 12 & 0.18&35 & 0.18& 1&  0.14& 1 & 0.14&9 & 0.19\\
15& 36&  0.16& 17 & 0.16&17 & 0.18& 20&  0.14& 17 & 0.14&11 & 0.17\\
16& 17&  0.16& 36 & 0.16&30 & 0.17& 31&  0.13& 34 & 0.13&15 & 0.16\\
17& 29&  0.12& 29 & 0.12&13 & 0.17& 28&  0.13& 31 & 0.13&17 & 0.13\\
18& 26&  0.12& 26 & 0.11&23 & 0.16& 34&  0.13& 28 & 0.13&16 & 0.11\\
19& 6&  0.12& 6 & 0.11&12 & 0.14& 4&  0.13& 4 & 0.13&4 & 0.10\\
20& 33&  0.11& 33 & 0.11&11 & 0.14& 14&  0.13& 11 & 0.13&18 & 0.09\\
21& 13&  0.11& 13 & 0.11&6 & 0.14& 2&  0.12& 2 & 0.13&25 & 0.09\\
22& 3&  0.10& 3 & 0.10&26 & 0.13& 5&  0.12& 5 & 0.12&13 & 0.09\\
23& 16&  0.10& 16 & 0.10&16 & 0.13& 3&  0.11& 18 & 0.12&19 & 0.08\\
24& 35&  0.08& 35 & 0.09&29 & 0.12& 11&  0.11& 3 & 0.11&1 & 0.07\\
25& 22&  0.08& 22 & 0.08&22 & 0.11& 6&  0.10& 30 & 0.11&31 & 0.06\\
26& 5&  0.08& 5 & 0.08&25 & 0.11& 21&  0.10& 24 & 0.10&23 & 0.06\\
27& 32&  0.07& 32 & 0.08&32 & 0.11& 24&  0.10& 14 & 0.09&30 & 0.05\\
28& 2&  0.07& 10 & 0.07&2 & 0.10& 30&  0.09& 21 & 0.08&34 & 0.03\\
29& 10&  0.07& 2 & 0.07&3 & 0.10& 12&  0.08& 35 & 0.08&28 & 0.03\\
30& 19&  0.07& 19 & 0.07&10 & 0.10& 35&  0.08& 12 & 0.07&22 & 0.03\\
31& 25&  0.07& 25 & 0.07&34 & 0.09& 33&  0.06& 33 & 0.07&8 & 0.02\\
32& 1&  0.07& 1 & 0.07&4 & 0.09& 18&  0.05& 6 & 0.06&32 & 0.02\\
33& 28&  0.06& 28 & 0.07&28 & 0.09& 15&  0.02& 36 & 0.03&29 & 0.02\\
34& 31&  0.06& 31 & 0.06&19 & 0.09& 27&  0.01& 15 & 0.03&26 & 0.01\\
35& 34&  0.06& 34 & 0.06&1 & 0.08& 9&  0.01& 9 & 0.01&10 & 0.01\\
36& 4&  0.06& 4 & 0.06&31 & 0.08& 36&  0.00& 27 & 0.00&12 & 0.01\\\hline
Rel\_Err & &  & & 0.028 & & 0.177  & & & & 0.045& & \revise{0.973} \\\hline
\end{tabular}}
\end{table}

\section{Conclusion and Future work}

In this paper, we present a framework to bridge the optimization and sensitivity analysis for computationally expensive functions via the adoption of the surrogate models. The optimization  and sensitivity analysis of the objective function $f(x)$ are performed sequently. The optimization procedure can be
 considered as an objective-oriented adaptive experimental design, where the objective is to fast catch the global shape of $f(x)$.  Its function evaluations are reused in order to help generate a faithful surrogate model $s(x)$, which is a lieu of $f(x)$ for the purpose of sensitivity analysis. Optimization, as  adaptive design, is showed to be a more effective experimental design method for the optimal LHD.  Furthermore,
we propose an objective-oriented adaptive experimental design method  for our proposed local multi-variate sensitivity analysis measures for high dimensional problems. It aims to make the generated response surface of better accuracy for the
calculation of the corresponding local sensitivity analysis measures than that based on the optimal LHD. We also demonstrate the advantage of Gaussian RBF over Kriging in cases of relatively high dimensionality and few experimental design points. In the future, we would like to extend the idea of objective-oriented   experimental design to other response surface assisted problems. 

\section{Acknowledgements}
This work was supported by  the 973 project No. 2015CB856000, the Natural Science Foundation of China, Grant
Nos. 11201054, 91330201 and by the Fundamental Research Funds for the Central Universities ZYGX2012J118, ZYGX2013Z005.

\bibliographystyle{acm}
\bibliography{references_SA,references_Surrogate_OPT,references_kriging,references_GO,references_DOE}

\end{document}